\documentclass[sigconf]{acmart}

\usepackage{eurosym}
\usepackage{comment}

\AtBeginDocument{%
  }

\setcopyright{acmlicensed}
\copyrightyear{}
\acmYear{}
\acmDOI{XXXXXXX.XXXXXXX}
\acmConference[]{}{}{}

\acmISBN{}

\begin{document}

\title{Static Is Not Enough: A Comparative Study of VR and SpaceMouse in Static and Dynamic Teleoperation Tasks}

\author{Yijun Zhou}
\affiliation{%
 \institution{Vrije Universiteit Amsterdam\\
Department of Computer Science
}
 \city{Amsterdam}
 \country{the Netherlands}}
\email{y.zhou7@student.vu.nl}

\author{Muhan Hou}
\affiliation{%
 \institution{Vrije Universiteit Amsterdam\\
Department of Computer Science
}
 \city{Amsterdam}
 \country{the Netherlands}}
\email{m.hou@vu.nl}

\author{Kim Baraka}
\affiliation{%
 \institution{Vrije Universiteit Amsterdam\\
Department of Computer Science
}
 \city{Amsterdam}
 \country{the Netherlands}}
\email{k.baraka@vu.nl}

\renewcommand{\shortauthors}{Zhou et al.}
\begin{abstract}
    
Imitation learning relies on high-quality demonstrations, and teleoperation is a primary way to collect them, making teleoperation interface choice crucial for the data. Prior work mainly focused on static tasks, i.e., discrete, segmented motions, yet demonstrations also include dynamic tasks requiring reactive control. As dynamic tasks impose fundamentally different interface demands, insights from static-task evaluations cannot generalize. To address this gap, we conduct a within-subjects study comparing a VR controller and a SpaceMouse across two static and two dynamic tasks ($N=25$). We assess success rate, task duration, cumulative success, alongside NASA-TLX, SUS, and open-ended feedback. Results show statistically significant advantages for VR: higher success rates, particularly on dynamic tasks, shorter successful execution times across tasks, and earlier successes across attempts, with significantly lower workload and higher usability. As existing VR teleoperation systems are rarely open-source or suited for dynamic tasks, we release our VR interface to fill this gap.

\end{abstract}



\maketitle

\section{INTRODUCTION}


Imitation learning is an effective approach for learning robot policies from user-provided data, and teleoperation is one of the most common ways to collect such demonstrations~\cite{belkhale2023data,10602544,mandlekar2022matters}. The choice of teleoperation interfaces therefore plays an important role in influencing the resulting data~\cite{hetrick2020comparing, ravichandar2020recent}, as it impacts how operators express their actions during control.

The demonstrations collected via teleoperation span a range of tasks, which can be grouped into static and dynamic tasks. Static tasks involve discrete, segmented motions, such as pick–and-place or reaching. Dynamic tasks, by contrast, require reactive, continuous control with time-sensitive adjustments~\cite{583093, 8280543}, such as flipping an object in a pan. As these two task types place fundamentally different demands on human control, an interface effective for static tasks may not necessarily perform well on dynamic ones.

Despite this distinction, prior evaluations of teleoperation interfaces have focused primarily on static tasks~\cite{li2025train, su2022mixed, mandlekar2018roboturk, chu2023bootstrapping}. As a result, existing insights do not directly clarify how well commonly used interfaces, such as VR controllers and SpaceMouse~\cite{li2025train, o2024open}, support dynamic tasks. In this work, VR refers solely to the use of a VR controller for 6-DoF tracking rather than a virtual environment. This gap makes it important to examine how different interfaces affect user task performance across both task types.

While several VR teleoperation systems have been used to collect high-quality demonstrations~\cite{mandlekar2018roboturk, su2022mixed, cuan2024leveraging, stotko2019vr}, many are not publicly available, limiting systematic comparison across interfaces. To address this, we introduce an open-source VR teleoperation interface and conduct a within-subjects comparison of VR and SpaceMouse across static and dynamic tasks. Our contributions are:

\begin{itemize}
    \item A systematic user study comparing VR and SpaceMouse (both 6 DoF) across static and dynamic tasks, evaluating user task performance and user experience.
    \item An open-source VR teleoperation interface that supports demonstration collection in dynamic settings.
\end{itemize}

\begin{figure}
    \centering
    \includegraphics[width=1\linewidth]{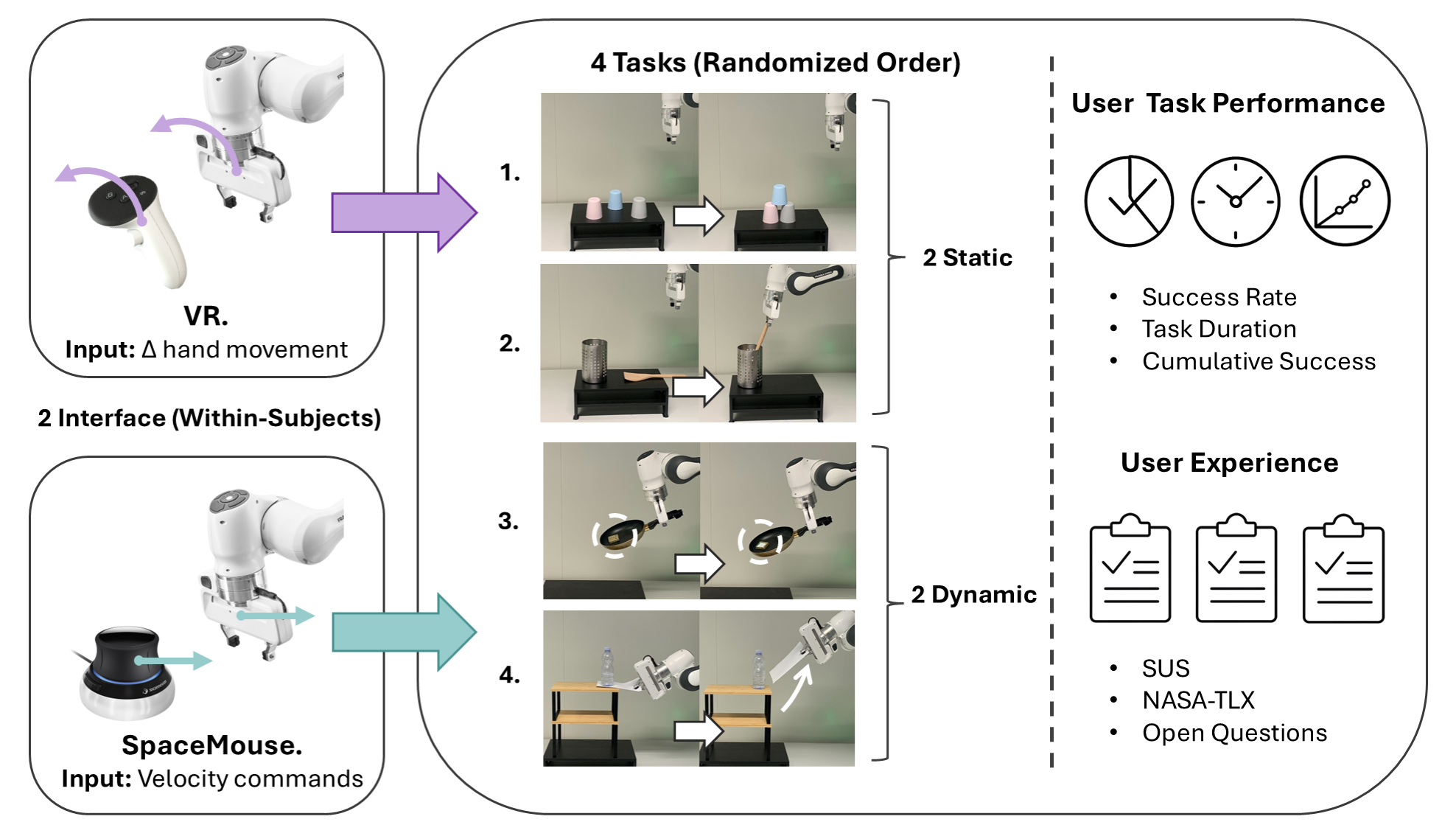}
    \caption{Overview of the study design, including the two teleoperation interfaces (VR, SpaceMouse), four randomized tasks (two static and two dynamic), and evaluated measures of user task performance and user experience.}
    \label{fig:overview}
\end{figure}

\section{RELATED WORK}

\subsection{Teleoperation Interface Comparisons}

VR controllers~\cite{mandlekar2018roboturk, su2022mixed, cuan2024leveraging, stotko2019vr} and SpaceMouse devices~\cite{chi2025diffusion, zhu2022bottom} have become the dominant teleoperation interfaces for collecting demonstrations in imitation learning. Prior comparisons across input devices~\cite{mandlekar2018roboturk, chu2023bootstrapping} and direct VR–SpaceMouse studies~\cite{li2025train, su2022mixed} report differences in usability, workload, and policy learning outcomes. However, these works focus almost exclusively on static or quasi-static manipulation, where motion is discrete and segmented.

This creates a key gap: static tasks cannot reveal how interfaces behave under reactive, continuous motion, nor how pose-based and velocity-based control differ when timing and momentum matter. Our work extends these comparisons to dynamic tasks, where these distinctions become clearer in how they influence user task performance and experience.

\subsection{Dynamic Manipulation and Data Collection}
Dynamic manipulation involves exploiting inertia and rapid motion rather than purely quasi-static control~\cite{583093}. Recent systems such as ALOHA~\cite{Zhao-RSS-23} and HiLSERL~\cite{luo2025precise} successfully collect high-quality demonstrations on dynamic tasks through teleoperation, yet they provide limited analysis of how interface choice influences user experience. Moreover, these systems typically rely on a single input device, which makes cross-interface comparisons difficult.

A separate line of work learns dynamic manipulation directly from human videos~\cite{chen2025tool, haldar2023teach, ha2022flingbot, zeng2020tossingbot}. These approaches bypass teleoperation altogether, but they lack real-time human feedback, leaving open questions about how different teleoperation interfaces support data collection for dynamic tasks.

Another direction studies dexterous in-hand manipulation~\cite{1249273,6907062, wang2025lessons, huang2023dynamic}, which centers on fine-grained finger motions rather than the larger, faster movements in arm-level dynamic tasks. Our study complements these efforts by examining interface effects in a more accessible setting.

\section{METHODOLOGY}
For VR teleoperation, we use the Oculus Quest~3 controller for 6-DoF tracking, while the headset is only used to capture motion and button inputs. Hand motion is mapped to robot end-effector commands via pose tracking~\cite{OrbikEbert2021OculusReader}. At each control cycle, the controller pose is transformed into the robot frame to obtain the hand rotation and position $(R^{hand}_t,\, P^{hand}_t)$.

Teleoperation is activated when the user presses any button, which records the current hand pose and the robot end-effector pose as references. End-effector motion is then generated from incremental hand motion. The positional command is computed from frame-to-frame displacement,
\begin{equation}
a^{pos}_t = \alpha \left(P^{hand}_t - P^{hand}_{t-1}\right),
\end{equation}
and applied to update the end-effector position. 

Rotational motion is obtained from the relative hand rotation,
\begin{equation}
a^{rot}_t =
\beta \cdot \mathrm{Log}\!\left(
R^{hand}_t (R^{hand}_{ref})^{-1}
\right),
\end{equation}
and applied through a quaternion update for numerical stability.

A calibration step is included to prevent VR drift: pressing the trigger resets the hand reference pose and its corresponding robot reference. This allows users to re-center their hand at any time without affecting the robot’s current pose.
\begin{figure}
    \centering
    \includegraphics[width=1\linewidth]{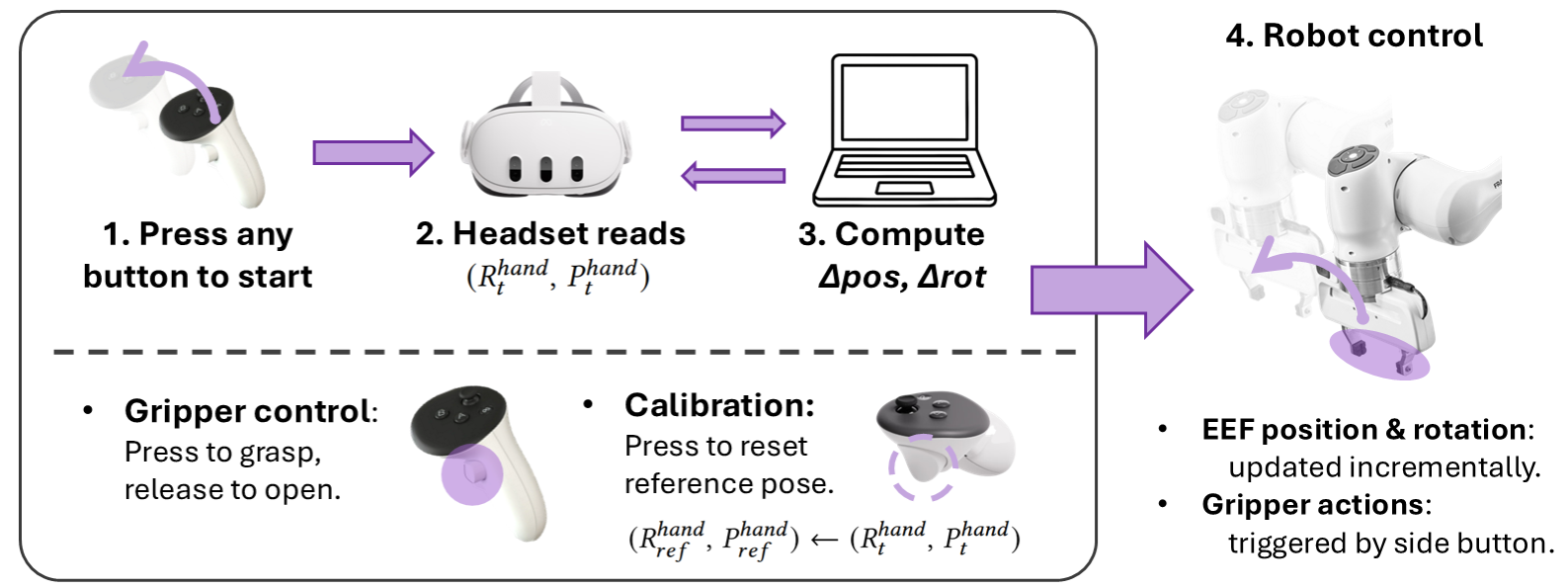}
    \caption{Workflow of our VR teleoperation interface. Controller pose is tracked and mapped to incremental end-effector commands, with button inputs handling gripper control and calibration.}
    \label{fig:vr_workflow}
\end{figure}

\section{EXPERIMENTAL SETUP}
\subsection{Ethics Statement}
The study was approved by the Research Ethics Review Committee of the Faculty of Science, Vrije Universiteit Amsterdam (Ref. 25-045).

\subsection{Hardware and Tasks}
We use a 7-DoF Franka Emika Panda robotic arm controlled via the Cartesian impedance controller in panda-py~\cite{elsner2023taming}. We use the Oculus Quest~3 controller for VR teleoperation. For SpaceMouse, we use a 3Dconnexion SpaceMouse Compact, and its six-axis velocity input is mapped to end-effector pose deltas via a scaling factor.

We choose stacking three cubes as the training task, which is a classic manipulation task that allows participants to practice the basic skills required by both interfaces: moving, rotating, grasping, and controller calibration (for VR).

\begin{figure}
    \centering
    \includegraphics[width=1\linewidth]{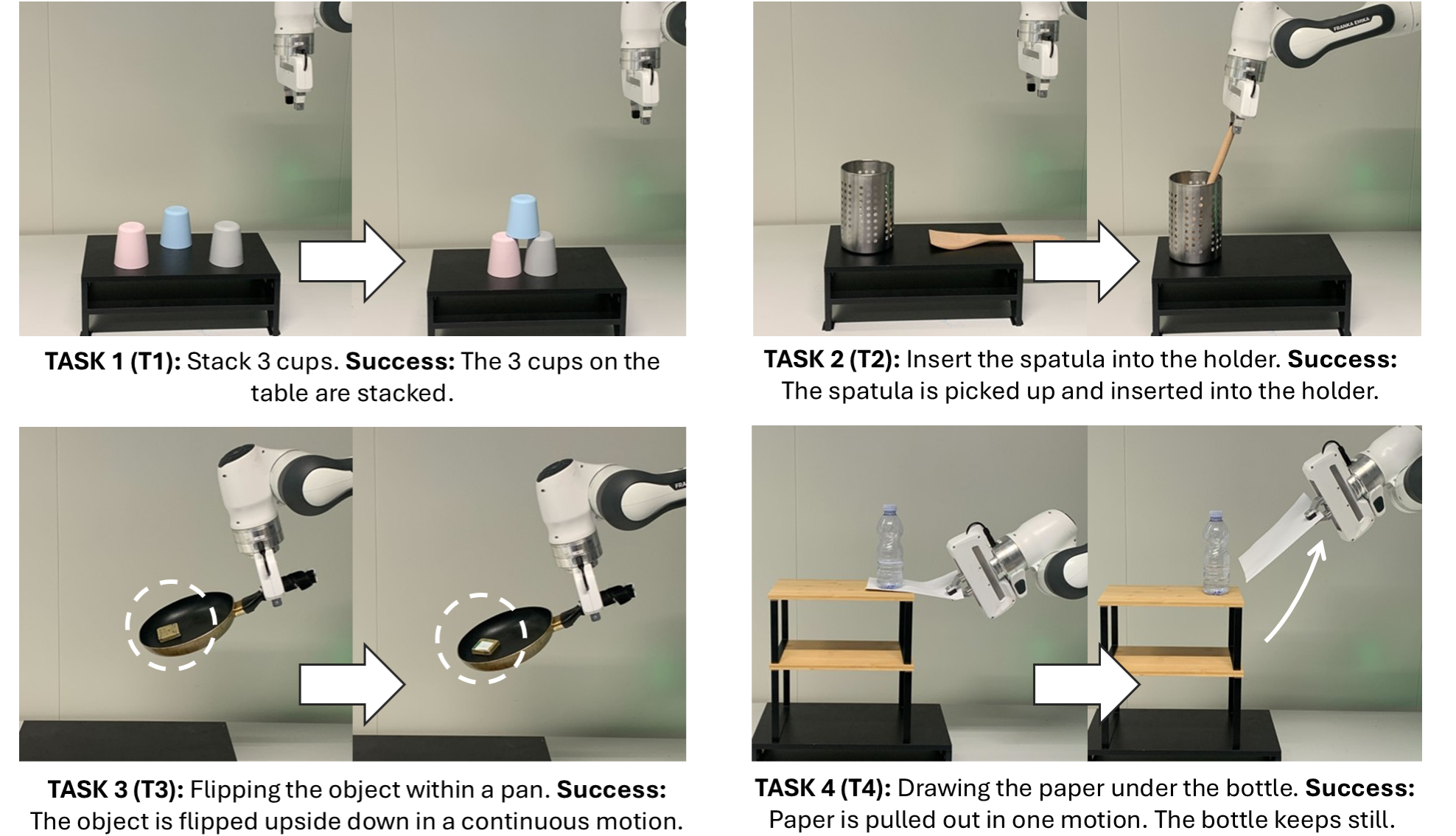}
\caption{4 Tasks used in our study, including success criteria. T1 and T2 are static tasks. T3 and T4 are dynamic tasks requiring continuous motion.}
    \label{fig:tasks}
\end{figure}

We select four tasks as our testbed (Figure~\ref{fig:tasks}), covering both static and dynamic tasks. For dynamic tasks, the action must be completed in a single continuous motion; slow or stop-and-go movements do not count as success. This structure allows us to compare user performance across task complexity and generalization.

The task order is in a constrained counterbalancing scheme. The two static tasks (T1, T2) and the two dynamic tasks (T3, T4) are randomly permuted separately. All participants complete the static pair before the dynamic pair, maintaining counterbalancing while providing a smooth transition from discrete to reactive actions.

\subsection{Participants}
We recruited 25 participants for a within-subjects study (17 male, 8 female, 0 other; mean age 24). Their prior experience with robots was mostly low, with 7 participants reporting no experience, 10 limited, 7 moderate, and 1 extensive. Prior gaming experience was generally higher, with 2 reporting no experience, 4 limited, 8 moderate, and 11 extensive. All participants completed both conditions.

Participants were recruited via on-campus posters. The experiment followed the ethical guidelines of our faculty’s research ethics board and and received its approval. Informed consent for participation and data collection was obtained before participation. Each participant received a \euro 10 gift card as compensation upon finishing.

\subsection{Procedure}
The experiment began with the VR controller condition. Participants first completed a training task that required one successful demonstration, with up to three minutes allowed. Participants who could not complete the task within this period were dismissed.

After training, participants performed four tasks. The order followed the constrained counterbalancing scheme: two static tasks followed by two dynamic ones, each internally randomized. For each task, participants had one minute to explore and up to five attempts, with each attempt capped at two minutes. A task ended once a single successful attempt was achieved. Attempts exceeded the time limit or failed the success criterion were recorded as failures. Object placements remained identical across all participants.

After the VR condition, participants filled out the NASA-TLX~\cite{HART1988139} and the System Usability Scale (SUS)~\cite{brooke1996quick}. After a one-minute break, they repeated the same procedure using the SpaceMouse interface. The fixed VR-SpaceMouse order was adopted for safety, as pilot observations showed that starting with VR would reduce early-stop issues by providing initial task familiarity.

At the end of the study, participants completed an open-ended questionnaire covering demographics, prior experience with robots and gaming, and subjective feedback.

\subsection{Metrics}
\subsubsection{User Task Performance}
We measure user task performance using success rate, task duration, and cumulative success. For each participant, we record the number of attempts until their first successful trial, whether they succeeded at all, and the duration of each demonstration. Success is labeled manually based on predefined criteria. The detailed definitions of these metrics are as follows:

\begin{itemize}
    \item \textbf{Success rate}: ratio of successful attempts to total attempts.
    \item \textbf{Task duration}: duration of the successful attempt, or the average duration across attempts for participants who do not succeed.
    \item \textbf{Cumulative success}: at each attempt number, the proportion of participants who have achieved at least one successful completion.
\end{itemize}

\subsubsection{User Experience}
User experience is assessed through workload, usability, and open-ended feedback. We use an unweighted version of NASA-TLX to measure mental demand, physical demand, temporal demand, perceived performance, effort, and frustration, on a 1–7 scale. Perceived usability is evaluated with the standard 10-item SUS, scored on a 0–100 scale. Participants also give open-ended feedback on interface preference, perceived strengths, difficulties, and suggestions for improvement.

\section{RESULTS AND DISCUSSION}
\subsection{User Task Performance}

\begin{figure}
    \centering
    \includegraphics[width=1\linewidth]{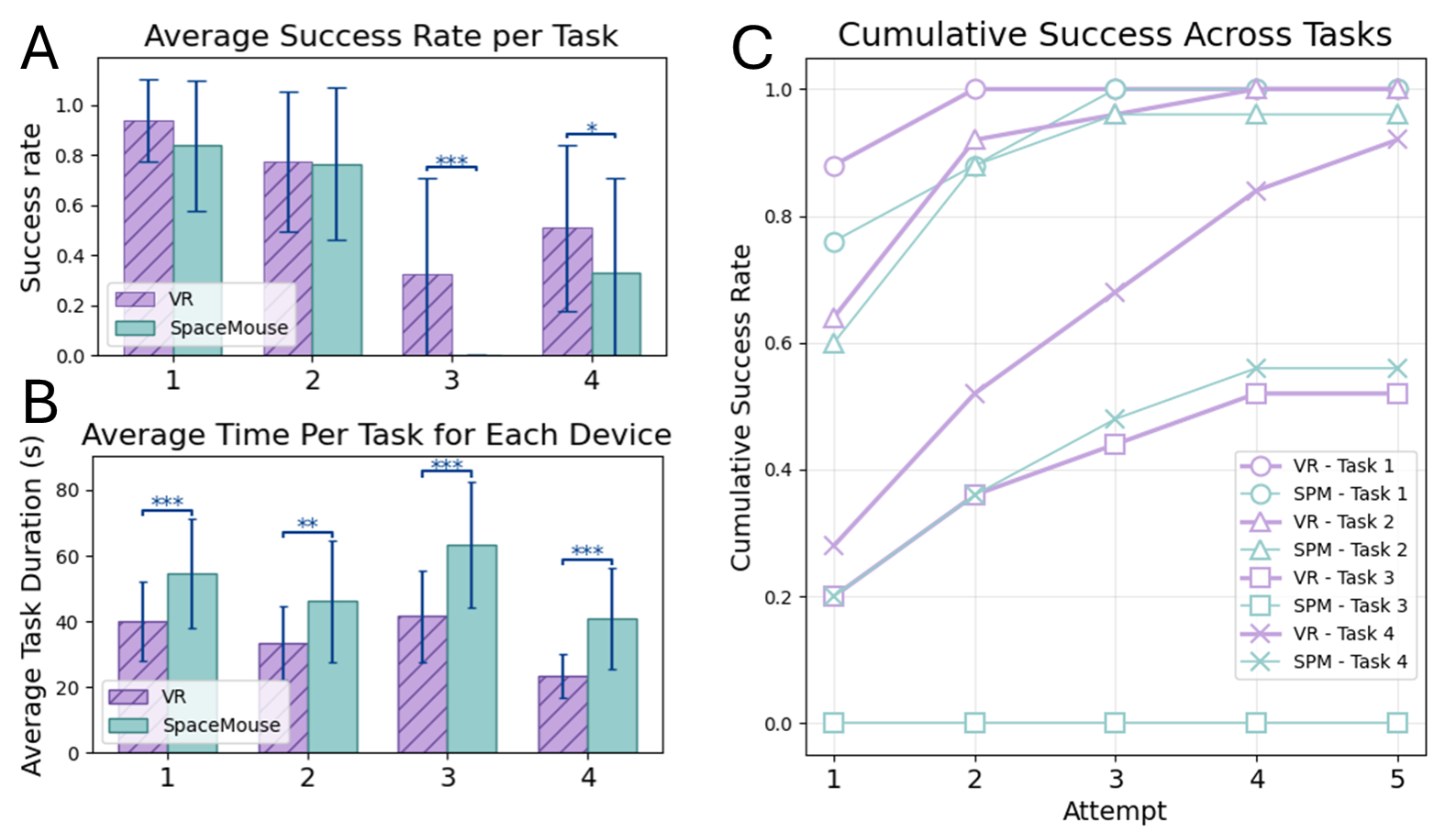}
    \caption{ User task performance across four tasks.
    A: Average success rate for VR and SpaceMouse across tasks.  
    B: Average completion time across tasks under both interfaces.  
    C: Cumulative success rate over the five allowed attempts for each task.  
    VR consistently achieves higher success rates, yields significantly shorter completion times, and reaches successful completion earlier across static and dynamic tasks.
    Significance levels are indicated as: \(*\,p < .05,\ **\,p < .01,\ ***\,p < .001\).
    }
    \label{fig:success+duration}
\end{figure}

Success rates decrease across tasks, with T3 being the most difficult (Figure~\ref{fig:success+duration}A). In the two static tasks, VR and SpaceMouse show no significant difference. Static manipulation relies on slow, incremental adjustments, and under such conditions both interfaces support similar success behavior (T1: 94.0\% vs. 84.0\%; T2: 77.7\% vs. 76.7\%).

Dynamic tasks present a sharply different pattern. Statistical tests show clear differences in both T3 and T4: VR outperforms SpaceMouse with a large gap in T3 (31.2\% vs. 0\%, $p < .001$) and a smaller but significant difference in T4 (50.9\% vs. 33.3\%, $p < .05$). The shift from static to dynamic tasks thus reveals where the two interfaces begin to diverge in success behavior.

T3 shows the strongest contrast. Flipping requires an impulsive wrist rotation that changes momentum abruptly. VR’s pose-delta input turns this motion into a sharp, high-acceleration command, while SpaceMouse outputs smoothed velocity, flattening impulsive input into uniform motion. This makes the required impulse hard to produce, explaining the low SpaceMouse success rate.

T4 results in low success for both interfaces, but for different reasons: participants often manage to pull the paper but trigger the robot’s velocity-limit safety via VR, while they frequently activate self-collision protection before reaching the paper with SpaceMouse. These distinct failure modes further demonstrate how the two interfaces behave differently under dynamic demands.

Task duration (Figure~\ref{fig:success+duration}B) shows significant differences across all tasks (T2: $p < .01$, T1/T3/T4: $p < .001$), with VR completing tasks consistently faster. These results indicate that once participants identify a solution, VR allows more direct execution, while SpaceMouse requires longer, segmented motion.

Cumulative success curves (Figure~\ref{fig:success+duration}C) suggest that our VR interface supports more accessible onboarding for users. For instance, T4 has a low overall success rate, yet 23 out of 25 participants eventually succeed by the fifth attempt via VR. This pattern highlights a clear trend: VR enables earlier success with fewer attempts, especially for dynamic tasks. 

\subsection{User Experience}

\begin{figure}
    \centering
    \includegraphics[width=1\linewidth]{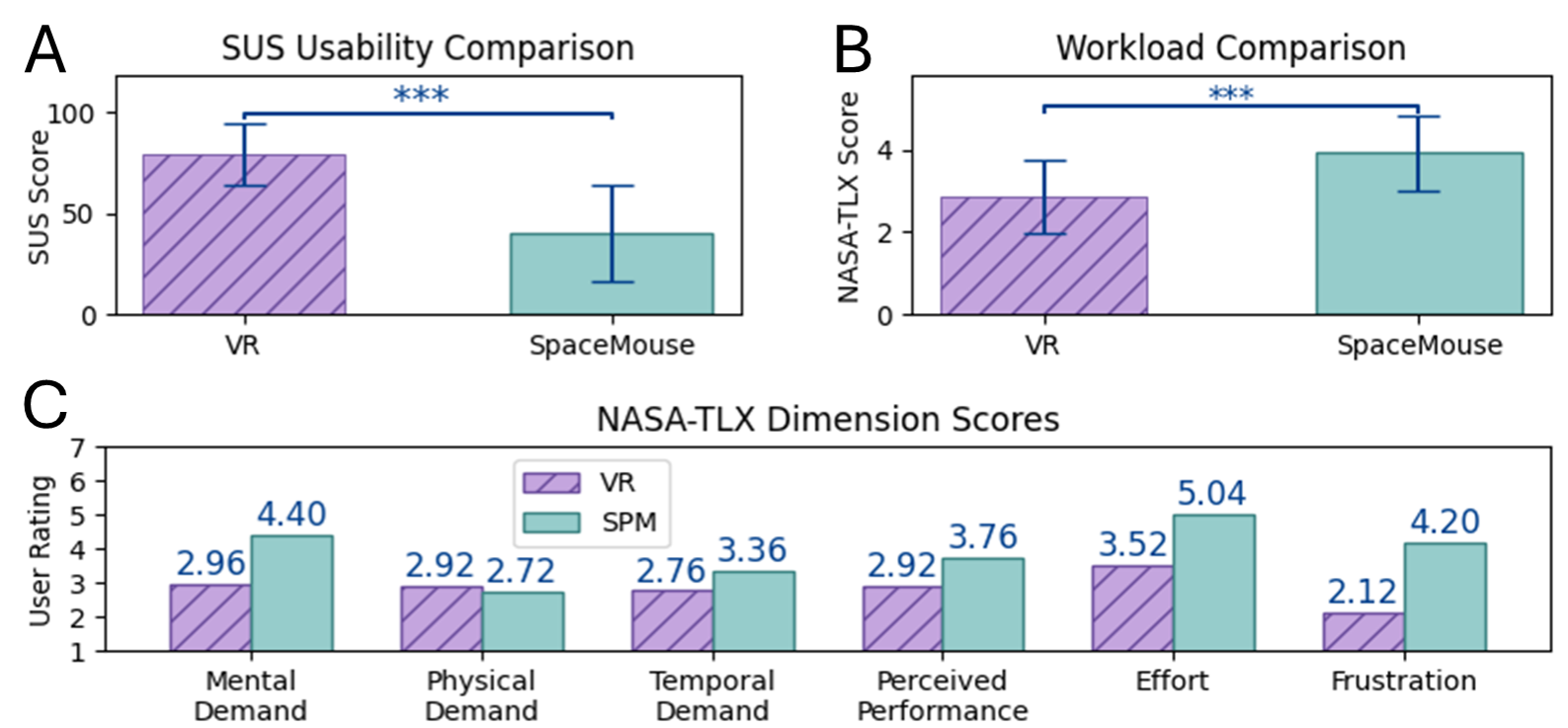}
    \caption{User subjective feedback.
    A: SUS scores. B: NASA-TLX scores. C: NASA-TLX scores for each individual dimension. VR shows higher usability and significantly lower workload across both static and dynamic tasks.}
    \label{fig:sus+nasa}
\end{figure}
The SUS results (Figure~\ref{fig:sus+nasa}A) show a clear difference between the two interfaces. VR receives a substantially higher usability score (M = 79.1) compared to SpaceMouse (M = 40.5). A paired t-test confirms that this difference is significant ($p < .001$). 

Similar patterns appear in the raw NASA-TLX scores (Figure~\ref{fig:sus+nasa}B, \ref{fig:sus+nasa}C). VR received lower workload ratings across all six dimensions, except for a slight increase in physical demand, likely due to the need to hold the VR controller throughout use. VR showed consistently lower perceived workload (M = 2.87) compared to SpaceMouse (M = 3.91), with the difference highly significant ($p < .001$).

For interface preference, 92\% participants preferred VR across tasks. They frequently described it as ``\textit{intuitive}". Many felt that its spatial mapping matched their hand movements, e.g.,

\begin{quote}
\textit{``VR felt more natural and intuitive. Like I was using my own hand."} -- P21
\end{quote}

A few positive comments were made about SpaceMouse: three participants liked its low physical effort and minimal calibration, consistent with its common use in teleoperation.

Participants also pointed out limitations for each interface. With the VR controller, three participants felt that impulsive hand movements transferred too much momentum to the robot, sometimes pushing it to its velocity limit. This reflects both the responsiveness of VR and its proneness to momentum-driven errors.

Feedback for SpaceMouse was more critical. Participants found its control logic unclear and often struggled to map intended motions to the robot, especially for rotation, e.g.,:
\begin{quote}
\textit{``The SpaceMouse is limited in how much it can tilt, rotate and move up and down."} - P13\\
\textit{``For spacemouse, rotation was very unpredictable."} - P5
\end{quote}
indicating the velocity-based input makes rotation cognitively heavier and less transparent to participants.

\subsection{Limitations and Future works}
This study has several limitations. Some participants noted that the transition from static to dynamic tasks was not fully smooth, and the difficulty gap may have influenced learning and performance. The study also has methodological constraints. The interfaces were not counterbalanced, as VR always preceded SpaceMouse, due to safety concerns. This fixed order may have introduced limited fatigue effects on the SpaceMouse results, despite the inclusion of a rest period. Additionally, trajectory data were not logged, which would prevent deeper analysis of movement quality, control strategies, and error patterns across interfaces.

Future work can address these issues by designing additional intermediate dynamic tasks to create a more gradual difficulty transition. This would allow a more thorough comparison between VR controllers and SpaceMouse on tasks with varying dynamism. Logging full end-effector and joint trajectories could also enable deeper analysis and support building a dataset for future research.

\section{CONCLUSION}
In this work, we study how teleoperation via a VR controller and a SpaceMouse differs in supporting demonstration collection for static and dynamic tasks. Results show that our VR interface yields higher success rates, shorter successful durations, and fewer attempts across all tasks, with participants reporting lower workload and higher usability. These findings indicate that pose-based control helps users produce fast or continuous motions, whereas the velocity-based SpaceMouse makes such actions harder to perform.

Overall, demonstrations collected with our VR interface are of higher quality, especially for dynamic tasks. As existing VR teleoperation systems are rarely open-source or suited for dynamic tasks, we release our VR interface to fill this gap. A promising direction for future work is to add end-effector trajectory logging and use the interface to build a dynamic-task dataset for future research.

\section*{CODE AND VIDEO}

Open-source code and documentation can be found at \url{https://github.com/ZoExOr/VR2Arm-Proj}. A video summarizing the contribution of the work and showing teleoperation examples with both interfaces is available through the supplementary material.

\bibliographystyle{ACM-Reference-Format}
\bibliography{000-referrence}

@inproceedings{mandlekar2022matters,
  title={What Matters in Learning from Offline Human Demonstrations for Robot Manipulation},
  author={Mandlekar, Ajay and Xu, Danfei and Wong, Josiah and Nasiriany, Soroush and Wang, Chen and Kulkarni, Rohun and Fei-Fei, Li and Savarese, Silvio and Zhu, Yuke and Mart{\'\i}n-Mart{\'\i}n, Roberto},
  booktitle={Conference on Robot Learning},
  pages={1678--1690},
  year={2022},
  organization={PMLR}
}

@article{ravichandar2020recent,
  title={Recent advances in robot learning from demonstration},
  author={Ravichandar, Harish and Polydoros, Athanasios S and Chernova, Sonia and Billard, Aude},
  journal={Annual review of control, robotics, and autonomous systems},
  volume={3},
  number={1},
  pages={297--330},
  year={2020},
  publisher={Annual Reviews}
}

@article{li2025train,
  title={How to Train Your Robots? The Impact of Demonstration Modality on Imitation Learning},
  author={Li, Haozhuo and Cui, Yuchen and Sadigh, Dorsa},
  journal={arXiv preprint arXiv:2503.07017},
  year={2025}
}

@inproceedings{mandlekar2018roboturk,
  title={Roboturk: A crowdsourcing platform for robotic skill learning through imitation},
  author={Mandlekar, Ajay and Zhu, Yuke and Garg, Animesh and Booher, Jonathan and Spero, Max and Tung, Albert and Gao, Julian and Emmons, John and Gupta, Anchit and Orbay, Emre and others},
  booktitle={Conference on Robot Learning},
  pages={879--893},
  year={2018},
  organization={PMLR}
}

@article{su2022mixed,
  title={Mixed-reality-enhanced human--robot interaction with an imitation-based mapping approach for intuitive teleoperation of a robotic arm-hand system},
  author={Su, Yun-Peng and Chen, Xiao-Qi and Zhou, Tony and Pretty, Christopher and Chase, Geoffrey},
  journal={Applied Sciences},
  volume={12},
  number={9},
  pages={4740},
  year={2022},
  publisher={MDPI}
}

@INPROCEEDINGS{Zhao-RSS-23, 
    AUTHOR    = {Tony Z. Zhao AND Vikash Kumar AND Sergey Levine AND Chelsea Finn}, 
    TITLE     = {{Learning Fine-Grained Bimanual Manipulation with Low-Cost Hardware}}, 
    BOOKTITLE = {Proceedings of Robotics: Science and Systems}, 
    YEAR      = {2023}, 
    ADDRESS   = {Daegu, Republic of Korea}, 
    MONTH     = {July}, 
    DOI       = {10.15607/RSS.2023.XIX.016} 
}

@inproceedings{hetrick2020comparing,
  title={Comparing virtual reality interfaces for the teleoperation of robots},
  author={Hetrick, Rebecca and Amerson, Nicholas and Kim, Boyoung and Rosen, Eric and de Visser, Ewart J and Phillips, Elizabeth},
  booktitle={2020 Systems and Information Engineering Design Symposium (SIEDS)},
  pages={1--7},
  year={2020},
  organization={IEEE}
}

@article{belkhale2023data,
  title={Data quality in imitation learning},
  author={Belkhale, Suneel and Cui, Yuchen and Sadigh, Dorsa},
  journal={Advances in neural information processing systems},
  volume={36},
  pages={80375--80395},
  year={2023}
}

@article{cuan2024leveraging,
  title={Leveraging haptic feedback to improve data quality and quantity for deep imitation learning models},
  author={Cuan, Catie and Okamura, Allison and Khansari, Mohi},
  journal={IEEE Transactions on Haptics},
  volume={17},
  number={4},
  pages={984--991},
  year={2024},
  publisher={IEEE}
}

@inproceedings{stotko2019vr,
  title={A VR system for immersive teleoperation and live exploration with a mobile robot},
  author={Stotko, Patrick and Krumpen, Stefan and Schwarz, Max and Lenz, Christian and Behnke, Sven and Klein, Reinhard and Weinmann, Michael},
  booktitle={2019 IEEE/RSJ International Conference on Intelligent Robots and Systems (IROS)},
  pages={3630--3637},
  year={2019},
  organization={IEEE}
}

@inproceedings{chu2023bootstrapping,
  title={Bootstrapping robotic skill learning with intuitive teleoperation: Initial feasibility study},
  author={Chu, Xiangyu and Tang, Yunxi and Kwok, Lam Him and Cai, Yuanpei and Au, Kwok Wai Samuel},
  booktitle={International Symposium on Experimental Robotics},
  pages={42--52},
  year={2023},
  organization={Springer}
}

@inproceedings{ha2022flingbot,
  title={Flingbot: The unreasonable effectiveness of dynamic manipulation for cloth unfolding},
  author={Ha, Huy and Song, Shuran},
  booktitle={Conference on Robot Learning},
  pages={24--33},
  year={2022},
  organization={PMLR}
}

@INPROCEEDINGS{583093,
  author={Mason, M.T. and Lynch, K.M.},
  booktitle={Proceedings of 1993 IEEE/RSJ International Conference on Intelligent Robots and Systems (IROS '93)}, 
  title={Dynamic manipulation}, 
  year={1993},
  volume={1},
  number={},
  pages={152-159 vol.1},
  doi={10.1109/IROS.1993.583093}}

@ARTICLE{8280543,
  author={Ruggiero, Fabio and Lippiello, Vincenzo and Siciliano, Bruno},
  journal={IEEE Robotics and Automation Letters}, 
  title={Nonprehensile Dynamic Manipulation: A Survey}, 
  year={2018},
  volume={3},
  number={3},
  pages={1711-1718},
  doi={10.1109/LRA.2018.2801939}}

@inproceedings{chen2025tool,
  title={Tool-as-Interface: Learning Robot Policies from Observing Human Tool Use},
  author={Chen, Haonan and Zhu, Cheng and Liu, Shuijing and Li, Yunzhu and Driggs-Campbell, Katherine Rose},
  booktitle={Proceedings of Robotics: Conference on Robot Learning (CoRL)},
  year={2025}
}

@inproceedings{wang2025lessons,
  title={Lessons from Learning to Spin “Pens”},
  author={Wang, Jun and Yuan, Ying and Che, Haichuan and Qi, Haozhi and Ma, Yi and Malik, Jitendra and Wang, Xiaolong},
  booktitle={Conference on Robot Learning},
  pages={3124--3138},
  year={2025},
  organization={PMLR}
}

@inproceedings{haldar2023teach,
  title={Teach a Robot to FISH: Versatile Imitation from One Minute of Demonstrations},
  author={Haldar, Siddhant and Pari, Jyothish and Rai, Anant and Pinto, Lerrel},
  booktitle={Robotics: Science and Systems},
  year={2023}
}

@article{zeng2020tossingbot,
  title={Tossingbot: Learning to throw arbitrary objects with residual physics},
  author={Zeng, Andy and Song, Shuran and Lee, Johnny and Rodriguez, Alberto and Funkhouser, Thomas},
  journal={IEEE Transactions on Robotics},
  volume={36},
  number={4},
  pages={1307--1319},
  year={2020},
  publisher={IEEE}
}

@inproceedings{huang2023dynamic,
  title={Dynamic Handover: Throw and Catch with Bimanual Hands},
  author={Huang, Binghao and Chen, Yuanpei and Wang, Tianyu and Qin, Yuzhe and Yang, Yaodong and Atanasov, Nikolay and Wang, Xiaolong},
  booktitle={Conference on Robot Learning},
  pages={1887--1902},
  year={2023},
  organization={PMLR}
}

@article{luo2025precise,
  title={Precise and dexterous robotic manipulation via human-in-the-loop reinforcement learning},
  author={Luo, Jianlan and Xu, Charles and Wu, Jeffrey and Levine, Sergey},
  journal={Science Robotics},
  volume={10},
  number={105},
  pages={eads5033},
  year={2025},
  publisher={American Association for the Advancement of Science}
}

@INPROCEEDINGS{1249273,
  author={Namiki, A. and Imai, Y. and Ishikawa, M. and Kaneko, M.},
  booktitle={Proceedings 2003 IEEE/RSJ International Conference on Intelligent Robots and Systems (IROS 2003) (Cat. No.03CH37453)}, 
  title={Development of a high-speed multifingered hand system and its application to catching}, 
  year={2003},
  volume={3},
  number={},
  pages={2666-2671 vol.3},
  doi={10.1109/IROS.2003.1249273}}

@article{zhu2022bottom,
  title={Bottom-up skill discovery from unsegmented demonstrations for long-horizon robot manipulation},
  author={Zhu, Yifeng and Stone, Peter and Zhu, Yuke},
  journal={IEEE Robotics and Automation Letters},
  volume={7},
  number={2},
  pages={4126--4133},
  year={2022},
  publisher={IEEE}
}

@article{chi2025diffusion,
  title={Diffusion policy: Visuomotor policy learning via action diffusion},
  author={Chi, Cheng and Xu, Zhenjia and Feng, Siyuan and Cousineau, Eric and Du, Yilun and Burchfiel, Benjamin and Tedrake, Russ and Song, Shuran},
  journal={The International Journal of Robotics Research},
  volume={44},
  number={10-11},
  pages={1684--1704},
  year={2025},
  publisher={Sage Publications Sage UK: London, England}
}

@INPROCEEDINGS{6907062,
  author={Dafle, Nikhil Chavan and Rodriguez, Alberto and Paolini, Robert and Tang, Bowei and Srinivasa, Siddhartha S. and Erdmann, Michael and Mason, Matthew T. and Lundberg, Ivan and Staab, Harald and Fuhlbrigge, Thomas},
  booktitle={2014 IEEE International Conference on Robotics and Automation (ICRA)}, 
  title={Extrinsic dexterity: In-hand manipulation with external forces}, 
  year={2014},
  volume={},
  number={},
  pages={1578-1585},
  doi={10.1109/ICRA.2014.6907062}}

@ARTICLE{10602544,
  author={Zare, Maryam and Kebria, Parham M. and Khosravi, Abbas and Nahavandi, Saeid},
  journal={IEEE Transactions on Cybernetics}, 
  title={A Survey of Imitation Learning: Algorithms, Recent Developments, and Challenges}, 
  year={2024},
  volume={54},
  number={12},
  pages={7173-7186},
  doi={10.1109/TCYB.2024.3395626}}

@inproceedings{o2024open,
  title={Open x-embodiment: Robotic learning datasets and rt-x models: Open x-embodiment collaboration 0},
  author={O’Neill, Abby and Rehman, Abdul and Maddukuri, Abhiram and Gupta, Abhishek and Padalkar, Abhishek and Lee, Abraham and Pooley, Acorn and Gupta, Agrim and Mandlekar, Ajay and Jain, Ajinkya and others},
  booktitle={2024 IEEE International Conference on Robotics and Automation (ICRA)},
  pages={6892--6903},
  year={2024},
  organization={IEEE}
}

@incollection{HART1988139,
title = {Development of NASA-TLX (Task Load Index): Results of Empirical and Theoretical Research},
editor = {Peter A. Hancock and Najmedin Meshkati},
series = {Advances in Psychology},
publisher = {North-Holland},
volume = {52},
pages = {139-183},
year = {1988},
booktitle = {Human Mental Workload},
issn = {0166-4115},
doi = {https://doi.org/10.1016/S0166-4115(08)62386-9},
url = {https://www.sciencedirect.com/science/article/pii/S0166411508623869},
author = {Sandra G. Hart and Lowell E. Staveland}
}

@book{brooke1996quick,
  author = {Brooke, John},
  month = {June},
  note = {ISBN: 9780748404605},
  publisher = {CRC Press},
  title = {"SUS-A quick and dirty usability scale." Usability evaluation in industry},
  url = {https://www.crcpress.com/product/isbn/9780748404605},
  year = 1996
}

@article{elsner2023taming,
title = {Taming the Panda with Python: A powerful duo for seamless robotics programming and integration},
journal = {SoftwareX},
volume = {24},
pages = {101532},
year = {2023},
issn = {2352-7110},
doi = {https://doi.org/10.1016/j.softx.2023.101532},
url = {https://www.sciencedirect.com/science/article/pii/S2352711023002285},
author = {Jean Elsner}
}

@misc{OrbikEbert2021OculusReader,
  author = {Jedrzej Orbik, Frederik Ebert},
  title = {Oculus Reader: Robotic Teleoperation Interface},
  year = {2021},
  url = {https://github.com/rail-berkeley/oculus_reader},
  note = {Accessed: YYYY-MM-DD}
}
\end{document}